\documentclass[11pt,a4paper]{article}
\usepackage{emnlp2021}
\usepackage{times}
\usepackage{latexsym}

\usepackage{microtype}

\usepackage{subcaption}

\usepackage{multirow}
\usepackage{amsmath}
\usepackage{capt-of}
\usepackage{tabularx}
\usepackage{epsfig}
\usepackage{amssymb}
\usepackage{amsfonts}
\usepackage{booktabs}
\usepackage{scalerel}
\usepackage[inline]{enumitem}
\usepackage{listings}
\usepackage{varwidth}
\usepackage[export]{adjustbox}
\usepackage{tikz}
\usetikzlibrary{tikzmark}
\usepackage{cleveref}

\usepackage{stmaryrd}
\usepackage{bbm}

\usepackage{algorithm}
\usepackage[noend]{algpseudocode}

\definecolor{deepblue}{rgb}{0,0,0.5}
\definecolor{officeblue}{RGB}{0,102,204}
\definecolor{deepred}{rgb}{0.6,0,0}
\definecolor{deepgreen}{rgb}{0,0.5,0}
\definecolor{mybrickred}{RGB}{182,50,28}

\definecolor{fillcolor}{RGB}{216,217,252}

\algnewcommand\algorithmicrequireb{{\hspace{0.85cm}}}
\algnewcommand\INPTDESCB{\item[\algorithmicrequireb]}

\algnewcommand\algorithmicfuncdesc{\textbf{Function:}}
\algnewcommand\FUNCDESC{\item[\algorithmicfuncdesc]}
\algnewcommand\algorithmicfuncdescb{{\hspace{1.48cm}}}
\algnewcommand\FUNCDESCB{\item[\algorithmicfuncdescb]}
\algnewcommand{\algorithmicgoto}{\textbf{goto}}
\algnewcommand{\Goto}[1]{\algorithmicgoto~\ref{#1}}

\usepackage{amsmath,amsfonts,bm}

\def\eqref#1{equation~\ref{#1}}

\def\1{\bm{1}}

\DeclareMathAlphabet{\mathsfit}{\encodingdefault}{\sfdefault}{m}{sl}
\SetMathAlphabet{\mathsfit}{bold}{\encodingdefault}{\sfdefault}{bx}{n}

\newcommand{\Ls}{\mathcal{L}}

\newcommand\our{\textsc{mT6}}
\newcommand\mtfive{\textsc{mT5}}

\usepackage{pifont}
\newcommand{\cmark}{{\color{blue}\ding{51}}}%
\newcommand{\xmark}{{\color{red}\ding{55}}}%

\title{mT6: Multilingual Pretrained Text-to-Text Transformer \\ with Translation Pairs}

\author{Zewen Chi$^{\dag\ddag}$\thanks{\ \  Contribution during internship at Microsoft Research.},~~Li Dong$^\ddag$,~~Shuming Ma$^{\ddag}$,~~Shaohan Huang$^{\ddag}$\\
\textbf{Xian-Ling Mao}$^\dag$\textbf{,}~~\textbf{Heyan Huang}$^\dag$\textbf{,}~~\textbf{Furu Wei}$^\ddag$\\
$^\dag$Beijing Institute of Technology \\
$^\ddag$Microsoft Research \\
\texttt{\{czw,maoxl,hhy63\}@bit.edu.cn}
\\\texttt{\{lidong1,shumma,shaohanh,fuwei\}@microsoft.com} \\}

\date{}

\begin{document}
\maketitle
\begin{abstract}
Multilingual T5 (\mtfive{}; \citealt{mt5}) pretrains a sequence-to-sequence model on massive monolingual texts, which has shown promising results on many cross-lingual tasks.
In this paper, we improve \textbf{m}ultilingual \textbf{t}ext-\textbf{t}o-\textbf{t}ext \textbf{t}ransfer \textbf{T}ransformer with \textbf{t}ranslation pairs (\our{}).
Specifically, we explore three cross-lingual text-to-text pre-training tasks, namely, machine translation, translation pair span corruption, and translation span corruption.
In addition, we propose a partially non-autoregressive objective for text-to-text pre-training.
We evaluate the methods on eight multilingual benchmark datasets, including sentence classification, named entity recognition, question answering, and abstractive summarization.
Experimental results show that the proposed \our{} improves cross-lingual transferability over \mtfive{}.
\end{abstract}

\section{Introduction}
\label{sec:intro}

Multilingual pretrained language models, such as mBERT~\cite{bert}, have attracted increasing attention. They not only improve the performance on downstream multilingual NLP tasks~\cite{xlm,xlmr,mbart,xlme}, but also show an impressive cross-lingual transferability~\cite{wu2019beto,xlingual:mbert:iclr20,xtreme,infoxlm}.

Multilingual pretrained models are typically trained on multilingual unlabeled text with unsupervised language modeling tasks, e.g., masked language modeling~\cite{bert}, causal language modeling~\cite{xlm}, and span corruption~\cite{t5}. These unsupervised tasks are built upon large-scale monolingual texts.
In addition, several studies propose cross-lingual tasks that utilize translation data from multilingual parallel corpora, such as translation language modeling~\cite{xlm}, cross-lingual contrast~\cite{infoxlm}, and bidirectional word alignment~\cite{hu2020explicit}. Thanks to the translation data, the pretrained models produce better-aligned cross-lingual representations and obtain better cross-lingual transferability.

Recently, the multilingual text-to-text transfer Transformer (\mtfive{};~\citealt{mt5}) achieves state-of-the-art performance on several cross-lingual understanding benchmarks. \mtfive{} inherits the benefits of T5~\cite{t5} that treats every text processing problem as a text-to-text problem, i.e., the problem of generating some target text conditioned on the input text. Despite the effectiveness of \mtfive{}, how to improve \mtfive{} with translation data is still an open problem.

In this paper, we present \our{}, standing for improving multilingual text-to-text transfer Transformer with translation data. 
\our{} differs from \mtfive{} in terms of both pre-training tasks and the training objective.
We present three cross-lingual tasks for text-to-text Transformer pre-training, i.e., machine translation, translation pair span corruption, and translation span corruption. In the translation span corruption task, the model is trained to predict the text spans based on the input translation pair. The cross-lingual tasks encourage the model to align representations of different languages.
We also propose a new objective for text-to-text pre-training, called partially non-autoregressive (PNAT) decoding. The PNAT objective divides the target sequence into several groups, and constrains that the predictions should be only conditioned on the source tokens and the target tokens from the same group.

We conduct experiments on both multilingual understanding and generation tasks. Our \our{} model yields substantially better performance than \mtfive{} on eight benchmarks.
We also provide an empirical comparison of the cross-lingual pre-training tasks, where we evaluate several variants of \our{} under the same pre-training and fine-tuning procedure.
Moreover, our analysis indicates that the representations produced by \our{} are more cross-lingual transferable and better-aligned than \mtfive{}.

The contributions are summarized as follows:
\begin{itemize}
\item We introduce three cross-lingual tasks for text-to-text Transformer pre-training, which improves \mtfive{} with translation data.
\item We propose a partially non-autoregressive objective that pretrains the decoder to use more information from the source sequence.
\item We provide extensive evaluation results of various pre-training tasks and training objectives.
\end{itemize}

\section{Background on T5 and \mtfive{}}

Multilingual text-to-text transfer Transformer (\mtfive{};~\citealt{mt5}) is the multilingual variant of T5~\cite{t5} pretrained on the mC4~\cite{mt5} dataset, which consists of natural text in 101 languages drawn from the public Common Crawl web scrape.

\begin{figure}
\centering
\includegraphics[width=0.4\textwidth]{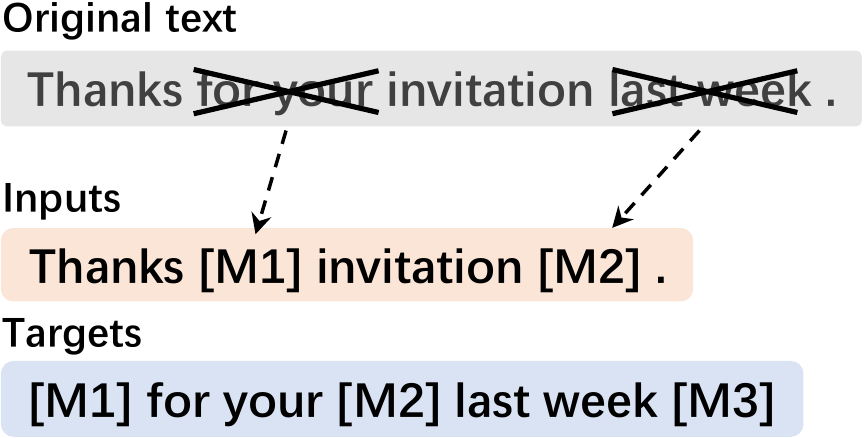}
\caption{Example of the span corruption task~\cite{t5} used in T5 and \mtfive{}.}
\label{fig:sc}
\end{figure}

The backbone architecture of \mtfive{} is the simple encoder-decoder Transformer~\cite{transformer}, which is trained in a unified text-to-text manner. In specific, text-based NLP problems are formulated as text-to-text transfer, i.e., the model is trained to predict the target text conditioned on the input source text. For example, in text classification, the model predicts the label text rather than a class index. This feature enables the \mtfive{} to be fine-tuned with the same training objective for every task.
Formally, let $x$ and $y$ denote the input sequence and the output sequence, the loss function of training the $x \rightarrow y$ transfer is
\begin{align}
\label{eq:sc}
\Ls(x \rightarrow y) = -\sum_{i = 1}^{|y|}{\log p(y_i | x, y_{<i})},
\end{align}
where $y_{<i} = y_1,\cdots,y_{i-1}$.
With the unified text-to-text formulation, the pre-training task can be designed by constructing the input and output text sequences. Specifically, \mtfive{} employs the span corruption task as the pre-training task, which is an unsupervised masked language modeling task.
As shown in Figure~\ref{fig:sc}, we provide an example of constructing the input and output sequences for span corruption. Given a natural sentence $s$, it first randomly selects several spans of $s$ as the spans to be masked.
Then, the input sequence is constructed by replacing the selected spans with unique mask tokens. The output sequence is the concatenation of the original tokens of the masked spans, each of which starts with a unique mask token to indicate the span to be decoded. We denote the above two operations as $g_i$ and $g_o$, standing for converting the original sentence $s$ into the input or the output formats of span corruption. Thus, the loss function of the span corruption task can be written as 
\begin{align}
\Ls_{\text{SC}}(s) = \Ls(g_i(s) \rightarrow g_o(s)).
\end{align}

\section{Methods}
\label{sec:method}

In this section, we first present three text-to-text pre-training tasks for improving \mtfive{} with translation data.
Then, we introduce the partially non-autoregressive decoding objective, and provide the detailed fine-tuning procedures for the classification, question answering, and named entity recognition tasks.

\begin{figure*}
\centering
\includegraphics[width=1.0\textwidth]{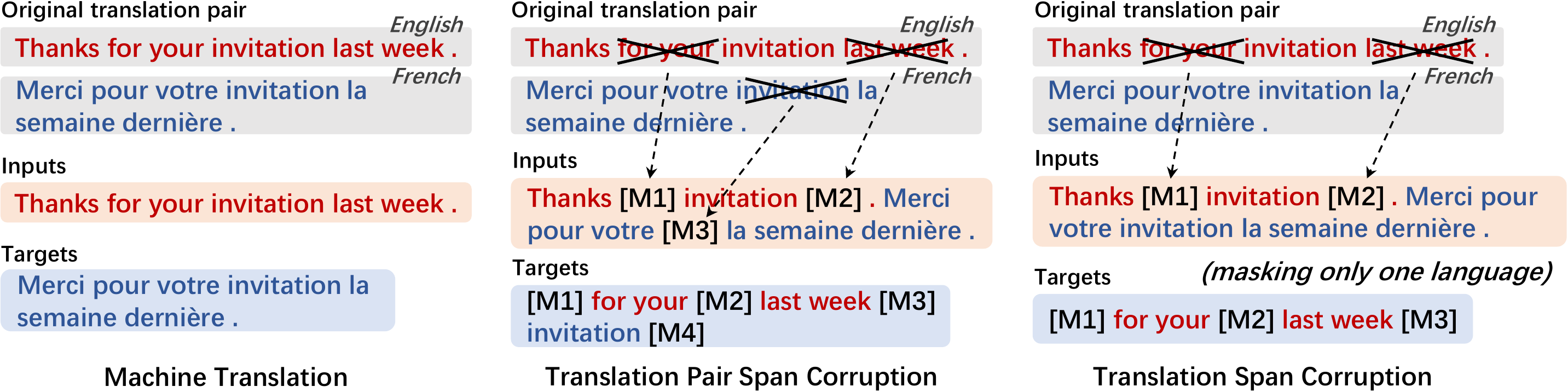}
\caption{Overview of three cross-lingual text-to-text pre-training tasks. For each task, we provide an example of the input and target text. The words marked with ``$\times$'' are randomly replaced with unique mask tokens like $\left[\text{M}_1\right]$. Notice that in the translation span corruption task, we mask tokens only in one language.}
\label{fig:tasks}
\end{figure*}

\subsection{Cross-lingual Pre-training Tasks with Translation Pairs}

As shown in Figure~\ref{fig:tasks}, we illustrate an overview of our cross-lingual text-to-text pre-training tasks. Given the same translation pair, the three tasks construct different input and output sequences. 

\subsubsection{Machine Translation}
\label{sec:mt}

Machine translation (MT) is a typical text-to-text task with the goal of translating a sentence from the source language into a target language.
It is a natural design to use MT as a text-to-text pre-training task for sequence-to-sequence learning~\cite{xnlg}.
Let $e$ and $f$ denote a sentence and its corresponding translation. We directly use $e$ and $f$ as the input and output sequences, respectively. The loss function of MT is
\begin{align}
\Ls_{\text{MT}}(e,f) = \Ls(e \rightarrow f).
\end{align}

\subsubsection{Translation Pair Span Corruption}
\label{sec:tpsc}

Inspired by the translation masked language modeling~\cite{xlm} task, we propose the translation pair span corruption (TPSC) task that aims to predict the masked spans from a translation pair instead of a monolingual sentence. 
Let $e$ and $f$ denote a sentence and its corresponding translation.
We concatenate $e$ and $f$ as a single sentence, and perform the span corruption on the concatenated sentence.
Formally, we construct the input and output sequences by $g_i([e;f])$ and $g_o([e;f])$, where $[e;f]$ stands for the concatenation of $e$ and $f$. With the resulting input and output sequences, the loss function of TPSC can be written as
\begin{align}
\Ls_{\text{TPSC}}(e,f) = \Ls(g_i([e;f]) \rightarrow g_o([e;f])).
\end{align}

\subsubsection{Translation Span Corruption}
\label{sec:tsc}

A potential issue of translation pair span corruption is that the spans in the target sequence can be organized in unnatural word order. As shown in Figure~\ref{fig:tasks}, the output sequence of TPSC is organized as ``\textit{$\left[\text{M}_1\right]$ for your $\left[\text{M}_2\right]$ last week $\left[\text{M}_3\right]$ invitation $\left[\text{M}_4\right]$}''. It can be found that the French word ``invitation'' is after the English word ``week'', which could harm the language model of the decoder. This motivates us to propose the translation span corruption (TSC) task where we only mask and predict the spans in one language.
Given a translation pair ($e$, $f$), we randomly select the $e$ or $f$ to perform span corruption. Without loss of generality, we consider $e$ as the sentence for span corruption. Then, the input and output sequences are constructed by $[g_i(e);f]$ and $g_o(e)$, respectively. With the resulting input and output sequences, the loss function of TSC can be written as
\begin{align}
\Ls_{\text{TSC}}(e,f) = \Ls([g_i(e);f]) \rightarrow g_o(e))).
\end{align}

\subsection{Pre-training Objective: Partially Non-autoregressive Decoding}

Recall that the predictions in \mtfive{} are conditioned on both the source tokens and the target tokens to the left. When predicting the tokens closer to the end, the model can use more information from the target sequence, resulting in the insufficient training of the encoder.

\begin{figure}
\centering
\includegraphics[width=0.48\textwidth]{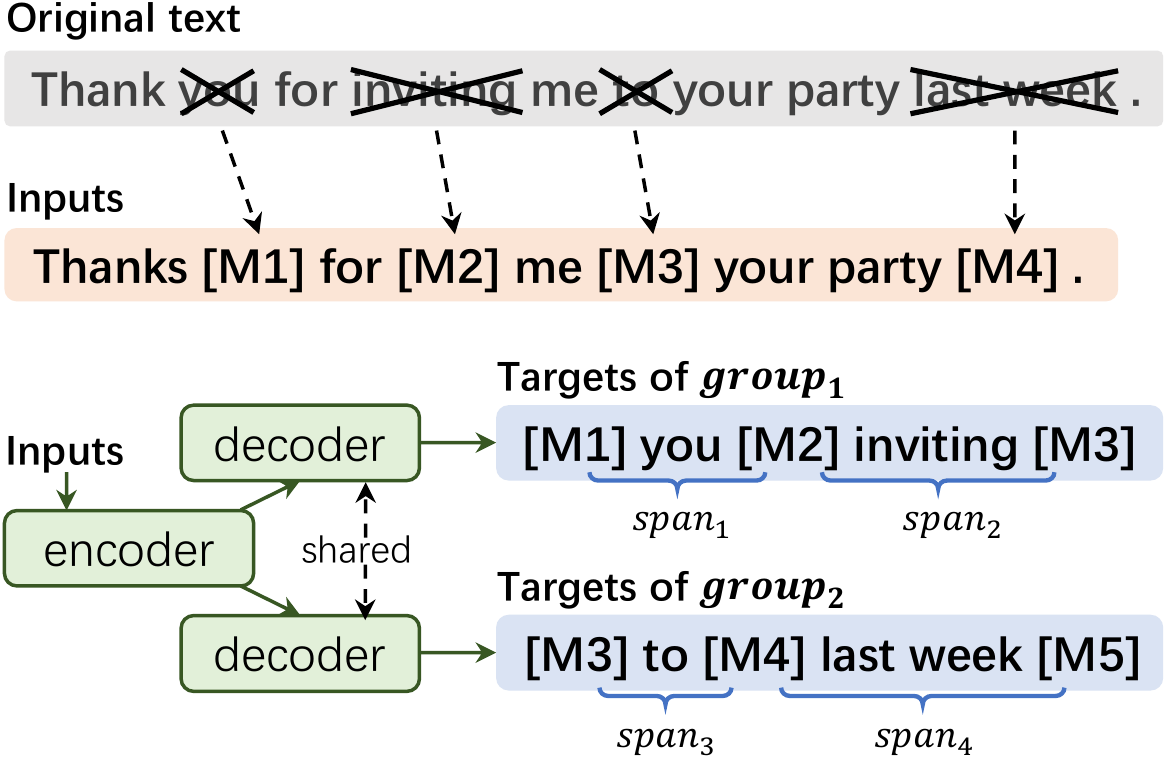}
\caption{Partially non-autoregressive objective.}
\label{fig:pnat}
\end{figure}

To encourage the model to utilize more information from the encoding side while preserving the ability of autoregressive decoding, we propose a new training objective for text-to-text training, called partially non-autoregressive decoding (PNAT). 
In Figure~\ref{fig:pnat}, we provide an example for PNAT.
Specifically, given a target sequence containing several spans, we divide the target sequence into groups, and train the model to decode each group separately. With the PNAT objective, a prediction is only conditioned on the source tokens and the target tokens from the same group.
Consider the target sequence consisting of $m$ spans. We divide the spans into $n_g$ groups, each of which contains $m/n_g$ consecutive spans. For the $j$-th group, we denote $l_j$ and $r_j$ as the start position and the end position, respectively.
The PNAT objective is defined as
\begin{align}
\Ls^{\textnormal{PNAT}}(x \rightarrow y) = -\sum_{j = 1}^{n_g}\sum_{i = l_j}^{r_j}{\log p(y_i | x, y_{l_j} \dots y_{i-1}}). \nonumber
\end{align}
The text-to-text loss $\Ls(x \rightarrow y)$ is a specially case of $\Ls^{\textnormal{PNAT}}(x \rightarrow y)$ with $n_g=1$.

The \our{} model is jointly pretrained on both monolingual and parallel corpora, where we use the span corruption and one of the three cross-lingual text-to-text tasks. For both tasks, we use the partially non-autoregressive decoding as the training objective where we divide the target sequence into $n_g$ groups. The overall pre-training objective is to minimize
\begin{align}
\Ls_{\our{}} = \Ls_{\text{SC}}^{\text{PNAT}}(s)+\Ls_{\text{X}}^{\text{PNAT}}(e,f),  &\\ \nonumber
X \in \{ \text{MT}, \text{TPSC}, \text{TSC} \}, &
\end{align}
where $\Ls_{\text{X}}^{\text{PNAT}}$ stands for the one of the loss functions of machine translation (MT; Section~\ref{sec:mt}), translation pair span corruption (TPSC; Section~\ref{sec:tpsc}) and translation span corruption (TSC; Section~\ref{sec:tsc}), with PNAT as the training objective.

\subsection{Cross-lingual Fine-tuning}

We fine-tune all parameters of the \our{} model with Equation~(\ref{eq:sc}) regardless of the end task.
Unlike language generation tasks, language understanding tasks should be pre-processed as the text-to-text format.
We introduce how to convert the following three types of the language understanding task into the text-to-text format, i.e., constructing the input and output sequences from the original examples.

\paragraph{Classification}

The goal of the text classification task is to predict the label of a given text. Following T5~\cite{t5}, we directly use the label text as the output text sequence. We provide an example for the MNLI natural language inference task~\cite{mnli2017}. Given an input sentence pair of ``\textit{You have access to the facts .}'' and ``\textit{The facts are accessible to you .}'', the goal is to classify the input into the relationships of ``\textit{entailment}'', ``\textit{contradiction}'', or ``\textit{neutral}''. The input and target sequences are constructed as

\textbf{Input:} \textit{$\left< bos \right>$ You have access to the facts. $\left< eos \right>$ The facts are accessible to you. $\left< eos \right>$ }

\textbf{Output:} \textit{$\left< bos \right>$ entailment $\left< eos \right>$}

Since multi-task fine-tuning is not the focus of this work, we do not prepend a task prefix in the input text. We also adopt a constrained decoding process, where the decoded text is constrained to be one of the labels.

\paragraph{Question Answering}

For the extractive question answering (QA) task, we concatenate the passage and the question as the input, and directly use the answer text as the target instead of predicting the answer span positions. We provide an example of converting a QA training example into the text-to-text format.

\textbf{Input:} \textit{$\left< bos \right>$ It has offices in Seoul, South Korea. $\left< eos \right>$ Where is the office in South Korea? $\left< eos \right>$ }

\textbf{Output:} \textit{$\left< bos \right>$ Seoul $\left< eos \right>$ }

We use the constrained decoding for the QA tasks
where we use the tokens shown in the input passage as the decoding vocabulary.

\paragraph{Named Entity Recognition}

In named entity recognition (NER), we do not directly use the original tag sentence as the output. We find that the model tends to repeat decoding the ``\textit{O}'' tag if the model directly learns to decode the tag sequences. Alternately, we construct the target text by concatenating the entity spans, each of which starts with the entity tag and ends with the entity tokens. We show an example of converting a NER training example into the text-to-text format.

\textbf{Input:} \textit{$\left< bos \right>$ Italy recalled Marcello Cuttitta . $\left< eos \right>$ }

\textbf{Output:} \textit{$\left< bos \right>$ $\left< loc \right>$ Italy $\left< sep \right>$ $\left< per \right>$ Marcello Cuttitta $\left< sep \right>$ $\left< eos \right>$ }

$\left< loc \right>$ and $\left< per \right>$ are entity tags denoting location and person. The $\left< sep \right>$ tag means the end of entity span. We use the following constrained decoding rules: (1) The model should decode entity tags or the end-of-sentence tag ($\left< eos \right>$) after a $\left< bos \right>$ token or a $\left< sep \right>$ token; (2) Otherwise, the model should decode the tokens from the input sentence or the $\left< sep \right>$ token for the other situations.

\section{Experiments}

\subsection{Setup}

\paragraph{Data}
Following previous work on cross-lingual pre-training~\cite{xlmr,infoxlm}, we use the natural sentences from CCNet~\cite{ccnet} in 94 languages for monolingual text-to-text tasks.
For cross-lingual text-to-text tasks, we use parallel corpora of 14 English-centric language pairs, collected from MultiUN~\cite{multiun}, IIT Bombay~\cite{iit}, OPUS~\cite{opus}, and WikiMatrix~\cite{wikimatrix}. Details of the pre-training data are described in Appendix.

\paragraph{Training Details}
In the experiments, we consider the small-size Transformer model~\cite{mt5}, with $d_\text{model}=512, d_\text{ff}=1,024$, $6$ attention heads, and $8$ layers for both the encoder and the decoder\footnote{Notice that the ``small-size'' defined in T5 and \mtfive{} are different. Here we follow the setting of \mtfive{}-small.}. We use the vocabulary provided by XLM-R~\cite{xlmr}, and extend it with $100$ unique mask tokens for the span corruption tasks. We pretrain our \our{} for $0.5$M steps with batches of $256$ length-$512$ input sequences. The model is optimized by the Adam optimizer~\cite{adam} with a linear learning rate scheduler. The pre-training procedure takes about $2.5$ days on an Nvidia DGX-2 Station. Details of the pre-training hyperparameters are described in Appendix.

\subsection{Results}

\begin{table*}[t]
\centering
\small
\scalebox{0.95}{
\begin{tabular}{c|ccccc|cccccc}
\toprule
\multirow{2}{*}{\bf Model} & \multicolumn{5}{c}{\bf Configuration}            & \bf Structured (F1) & \multicolumn{3}{c}{\bf Question Answering (F1)} & \multicolumn{2}{c}{\bf Classification (Acc.)}  \\
& SC & PNAT &MT&TPSC& TSC                 &    WikiAnn    & XQuAD & MLQA &           TyDiQA            & XNLI &             PAWS-X              \\ \midrule
NMT & \xmark&\xmark&\cmark&\xmark&\xmark                   &      27.3      & 12.5  & 14.9 &            16.8             & \bf {64.8} &              55.0            \\
\mtfive{} & \cmark&\xmark&\xmark&\xmark&\xmark              &      43.1      & 42.1  & 37.6 &            30.7             & 57.2 &              78.0               \\
\our{} (ours) & \cmark&\cmark&\xmark&\xmark&\cmark         &     \bf {44.7}      & \bf {50.4}  & \bf {44.1} &         \bf   {36.0}             &  {64.7} &            \bf  {82.2}               \\ \midrule
\multirow{4}{*}{Ablations} & \cmark&\cmark&\xmark&\xmark&\xmark           &      43.7      & 45.1  & 38.5 &            32.3             & 57.9 &              77.5               \\
& \cmark&\xmark&\cmark&\xmark&\xmark               &      43.9      & 38.5  & 33.3 &            29.4             & {65.9} &              79.3            \\
& \cmark&\xmark&\xmark&\cmark&\xmark     &      42.3      & 46.2  & 40.8 &            35.3             & 64.0 &              78.9            \\
& \cmark&\xmark&\xmark&\xmark&\cmark & 43.8 & 47.6 & 40.5 & {36.7} & 65.4 & 80.3 \\
\midrule
\midrule
\multicolumn{12}{l}{~~\textit{Pre-training with larger batch size and more training steps}} \\
\multicolumn{6}{l}{\mtfive{}~\cite{mt5}} & 50.5 & 58.1 & 54.6 & 35.2 & 67.5 & 82.4 \\
\bottomrule
\end{tabular}}
\caption{Evaluation results on XTREME under the cross-lingual transfer setting, where models are only fine-tuned on the English training data but evaluated on all target languages. We pretrain models with different combinations of span corruption (SC), machine translation (MT), translation pair span corruption (TPSC), translation span corruption (TSC), and partially non-autoregressive decoding (PNAT). All results are averaged over five runs.}
\label{table:ablation}
\end{table*}

\subsubsection{XTREME Cross-lingual Understanding}
To validate the performance of \our{}, we evaluate the pretrained models on XTREME~\cite{xtreme}, which is a widely used benchmark for cross-lingual understanding. Following \mtfive{}~\cite{mt5}, we consider six downstream tasks included by XTREME: 
the named entity recognition (NER) task on the WikiAnn~\cite{panx,rahimi2019} dataset in $40$ languages, the question answering (QA) task on MLQA~\cite{mlqa}, XQuAD~\cite{xquad}, and TyDiQA-GoldP~\cite{tydiqa}, the cross-lingual natural language inference task on XNLI~\cite{xnli}, and cross-lingual paraphrase adversaries on PAWS-X~\cite{pawsx}.
The models are evaluated under the cross-lingual transfer setting~\cite{xlmr,xtreme}. Under this setting, the models should be fine-tuned only on English training data but evaluated on all target languages.
Moreover, for each pretrained model, only one model is used for all languages rather than selecting fine-tuned models separately. Details of the fine-tuning hyperparameters are described in Appendix.

As shown in Table~\ref{table:ablation}, we present the evaluation results of the pretrained models on the XTREME benchmark. We observe that \our{} achieves the best performance on XTREME, improving the average score from $45.0$ to $50.4$, as we go from \mtfive{} to \our{}. 
It is worth mentioning that pre-training the model only with the machine translation task performs even worse than \mtfive{}. We have noticed that several target languages in TyDiQA and WikiAnn are not covered by our parallel corpora. However, the NMT pretrained model still shows poor results on the other four tasks, where all target languages are covered by the training data. Detailed results can be found in Appendix.

\subsubsection{Comparison of Pre-training Tasks}

To provide a clear comparison among the pre-training tasks, we implement the text-to-text pre-training methods presented in Section~\ref{sec:method}, and pretrain variants of \our{} with the same training data and resources for fair comparisons. 

Table~\ref{table:ablation} compares the evaluation results of the models pretrained with seven different combinations of span corruption (SC), machine translation (MT), translation pair span corruption (TPSC), translation span corruption (TSC), and partially non-autoregressive decoding (PNAT). It can be observed that jointly training SC+TSC with PNAT achieves the best overall performance on the XTREME benchmark, with substantial gains over the models trained on monolingual data only. The same trend can be observed for the other models pretrained on both monolingual data and parallel data.
This demonstrates that introducing translation data to text-to-text pre-training can improve the performance on the end tasks of cross-lingual understanding. Moreover, PNAT provides consistent gains over SC and SC+TSC, showing that PNAT is effective on both monolingual and cross-lingual tasks. Surprisingly, SC+PNAT obtains comparable results to SC+MT without any parallel data. Comparing TSC with MT and TPSC, we observe that SC+TSC brings noticeable improvements on question answering tasks. Although SC+MT shows competitive results on XNLI, the results on the other tasks are relatively low, indicating that simply jointly training SC with MT is not the most effective way to pretrain \our{}.

\begin{table*}[t]
\centering
\small
\scalebox{0.98}{
\begin{tabular}{llccc|ccc|ccc}
\toprule
\multirow{2}{*}{\bf Model} & \multirow{2}{*}{\bf \#Param} & \multicolumn{3}{c}{\bf en} & \multicolumn{3}{c}{\bf fr} & \multicolumn{3}{c}{\bf zh} \\
& & RG-1 & RG-2 & RG-L & RG-1 & RG-2 & RG-L & RG-1 & RG-2 & RG-L \\ \midrule
\multicolumn{11}{l}{~~\textit{Larger model size}} \\
XLM~\cite{xnlg} & 800M & 48.15 & 26.35 & 45.04 & 56.27 & 39.20 & 52.84 & 55.30 & 42.57 & 52.95  \\
XNLG~\cite{xnlg} & 800M & 48.76 & 26.82 & 45.57 & 57.84 & 40.81 & 54.24 & 57.65 & 44.93 & 54.95   \\ \midrule
\multicolumn{11}{l}{~~\textit{Our re-implementation (Fine-tuning with full training data)}} \\
\mtfive{} (reimpl) & 300M & 46.58  & 24.45  & 43.32  & 54.12  & 36.78  & 50.61  & 57.30  & 44.08  & 54.65   \\
\our{} & 300M & \textbf{46.82}  & \textbf{24.65}  & \textbf{43.50}  & \textbf{54.82}  & \textbf{37.61}  & \textbf{51.30}  & \textbf{57.38}  & \textbf{44.20}  & \textbf{54.66}   \\ \midrule
\multicolumn{11}{l}{~~ \textit{Our re-implementation (Fine-tuning with 1K training data)}} \\
\mtfive{} & 300M & 28.00  & 10.89  & 26.13  & 32.56  & 17.25  & 29.75  & 44.16  & 31.20  & 41.86 \\
\our{} & 300M & \textbf{28.80}  & \textbf{11.44}  & \textbf{26.45}  & \textbf{35.07}  & \textbf{18.70}  & \textbf{31.39}  & \textbf{46.48}  & \textbf{33.17}  & \textbf{44.02} \\
\bottomrule
\end{tabular}
}
\caption{Evaluation results on Gigaword multilingual abstractive summarization. RG is short for ROUGE. Results of XLM and XNLG are taken from~\cite{xnlg}. Results of \mtfive{} and \our{} are averaged over three runs.}
\label{table:giga}
\end{table*}

\subsection{Abstractive Summarization}

\paragraph{Multilingual Summarization}

In addition to language understanding tasks, we also evaluate our \our{} model on the abstractive summarization task. Abstractive summarization aims to generate a summary of the input document while preserving its original meaning. We use the Gigaword dataset provided by \citet{xnlg}. The dataset is constructed by extracting the first sentences and headlines as the input documents and summaries, respectively. The dataset consists of examples in the languages of English, French, and Chinese. For each language, it contains $500$K, $5$K, and $5$K examples for the training, validation, and test, respectively. We fine-tune the models for $20$ epochs with a batch size of $32$ and a learning rate of $0.00001$. During decoding, we use the greedy decoding for all evaluated models.

As shown in Table~\ref{table:giga}, we report the ROUGE~\cite{lin-2004-rouge} scores of the models on Gigaword multilingual abstractive summarization. We observe that \our{} consistently outperforms \mtfive{} on all the three target languages. Comparing with the XLM~\cite{xlm} and XNLG~\cite{xnlg} models with $800$M parameters, our \our{} model achieves a similar performance with only $300$M parameters. Besides, under the setting with fewer training data, \our{} shows more improvements over \mtfive{}.

\begin{table}[t]
\centering
\small
\begin{tabular}{lcccc}
\toprule
\bf Model & \bf es-en & \bf ru-en & \bf vi-en & \bf tr-en \\ \midrule
\mtfive{} & 11.36 & 8.77 & 8.98 & 10.57 \\
\our{}  & \bf 11.83 & \bf 9.49 & \bf 9.52 & \bf 10.80 \\
\bottomrule
\end{tabular}
\caption{ROUGE-2 scores on Wikilingua cross-lingual summarization. Results are averaged over three runs.}
\label{table:wikilingua}
\end{table}

\paragraph{Cross-Lingual Summarization}

The cross-lingual summarization task aims to generate summaries in a different language. We use the Wikilingua~\cite{wikilingua} dataset containing passage-summary pairs in four language pairs. We fine-tune the models for $100$K steps with a batch size of $32$ and a learning rate of $0.0001$. We use the greedy decoding for all evaluated models. The evaluation results are shown in Table~\ref{table:wikilingua}, where \our{} outperforms \mtfive{} on the test sets of four language pairs.

\begin{table}[t]
\centering
\small
\scalebox{0.98}{
\renewcommand\tabcolsep{3.3pt}
\begin{tabular}{lccccc}
\toprule
\bf Model & \textbf{XQuAD} & \textbf{MLQA} & \textbf{TyDiQA} & \bf XNLI & \bf PAWS-X \\ \midrule
\mtfive{} & 30.4 & 27.5 & 27.5 & 19.5 & 16.0 \\
\our{}  & \textbf{28.6} & \textbf{27.2} & \textbf{25.9} & \textbf{14.6} & \textbf{13.2}  \\
\bottomrule
\end{tabular}}
\caption{The cross-lingual transfer gap scores on the XTREME tasks. A lower transfer gap score indicates better cross-lingual transferability. We use the EM scores to compute the gap scores for the QA tasks.}
\label{table:gap}
\end{table}

\subsection{Cross-lingual Transfer Gap}
To explore whether our \our{} model achieves better cross-lingual transferability, we compare the cross-lingual transfer gap scores of our \our{} with \mtfive{}. Cross-lingual transfer gap~\cite{xtreme} is defined as the difference between the performance on the English test set and the average performance on the non-English test sets. The transfer gap indicates how much the end-task knowledge preserves when transferring from English to the other target languages. Empirically, a lower transfer gap score indicates better cross-lingual transferability. Following \citet{xtreme}, we compute the transfer gap scores over the sentence classification and question answering tasks.
As shown in Table~\ref{table:gap}, \our{} consistently reduces the transfer gap across all the five tasks, demonstrating that our model is more effective for cross-lingual transfer than \mtfive{}.

\begin{figure}
\centering
\includegraphics[width=0.49\textwidth]{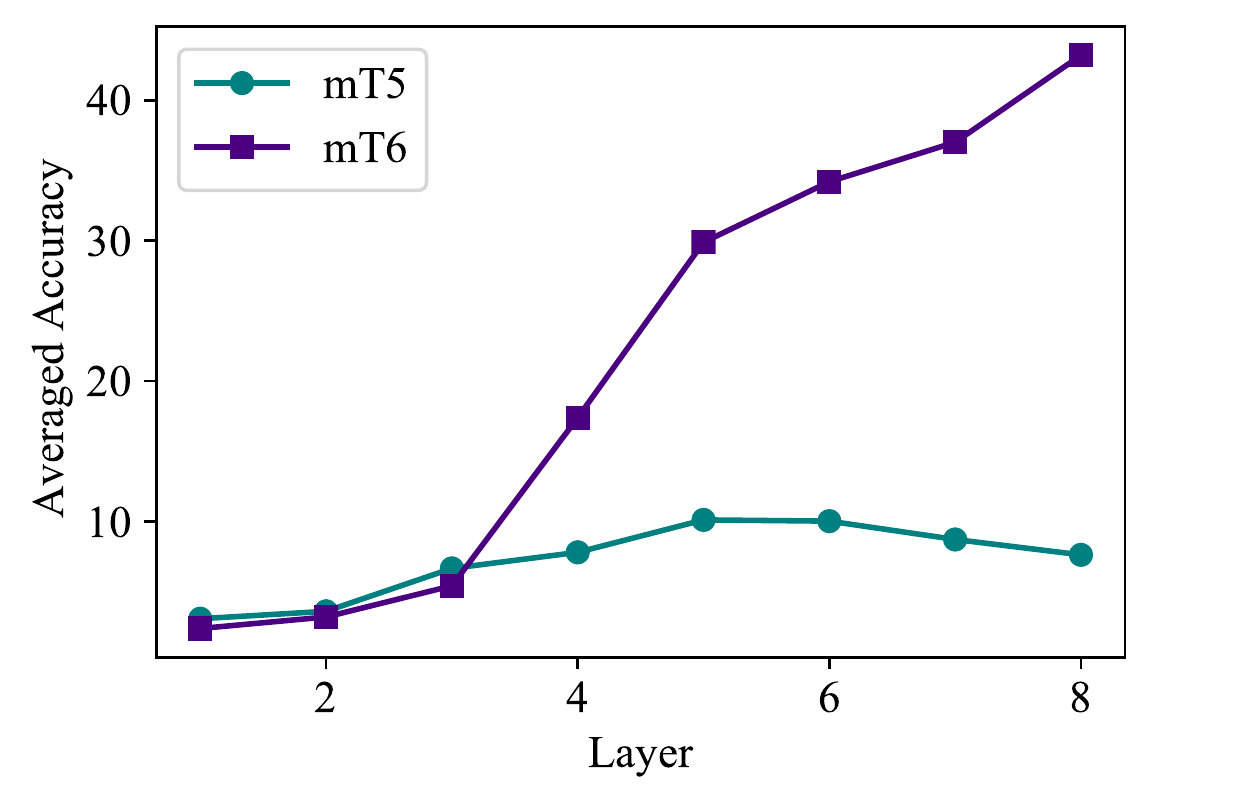}
\caption{Evaluation results of different layers on Tatoeba cross-lingual sentence retrieval. We illustrate the average accuracy@1 scores on the Tatoeba test sets of the 14 language pairs covered by the parallel data.}
\label{fig:xir}
\end{figure}

\subsection{Cross-lingual Representations}

We analyze the cross-lingual representations produced by our \our{} model. Following \citet{infoxlm}, we evaluate the representations on the Tatoeba~\cite{tatoeba} cross-lingual sentence retrieval task. The test sets consist of 14 English-centric language pairs covered by the parallel data in our experiments. Figure~\ref{fig:xir} illustrates the average accuracy@1 scores of cross-lingual sentence retrieval. The scores are averaged over 14 language pairs and both the directions of xx $\rightarrow$ en and en $\rightarrow$ xx.
From the figure, we observe that \mtfive{} shows a parabolic trend across different layers, which also appears in other cross-lingual encoder models~\cite{simalign,infoxlm}. Differently, we obtain better performance as we use higher layers of our \our{} model.
At layer-$8$, our \our{} model achieves an average accuracy@1 of $43.2$, outperforming the \mtfive{} model by $35.6$, which means our \our{} model produces better-aligned text representations. We believe the better-aligned representations potentially improve the cross-lingual transferability.
Furthermore, the results also indicate that our pre-training objective is more effective for training the encoder than \mtfive{}.

\begin{table}[t]
\centering
\scalebox{1}{
\begin{tabular}{lcccc}
\toprule
\bf Model & \bf en-de & \bf en-fr & \bf en-ro & \bf Avg \\ \midrule
\mtfive{} & 35.84 & 19.05 & 45.24 & 33.38 \\
\our{} & \bf 23.69 & \bf 12.11 & \bf 42.56 & \bf 26.12 \\
\bottomrule
\end{tabular}}
\caption{Evaluation results on word alignment. We report the alignment error rate scores (lower is better). We use the hidden vectors from the last encoder layer, and apply the SimAlign~\cite{simalign} tool to obtain the resulting word alignments.}
\label{table:wa}
\end{table}

\subsection{Word Alignment}

In addition to cross-lingual sentence retrieval that evaluates sentence-level representations, we also explore whether the representations produced by \our{} are better-aligned at token-level. Thus, we compare our \our{} with \mtfive{} on the word alignment task, where the goal is to find corresponding word pairs in a translation pair. We use the hidden vectors from the last encoder layer, and apply the SimAlign~\cite{simalign} tool to obtain the resulting word alignments. Table~\ref{table:wa} shows the alignment error rate (AER) scores on the test sets provided by \citet{simalign}. Among the three language pairs, \our{} achieves lower AER scores than \mtfive{}, indicating that the cross-lingual representations produced by \our{} are also better-aligned at token-level.

\begin{table}[t]
\centering
\small
\scalebox{1.0}{
\begin{tabular}{lcccc}
\toprule
\bf Noise Density & \textbf{NER} & \textbf{QA} & \textbf{Classification} & \bf Avg \\ \midrule
15\% & 41.7 & 33.5 & 71.9 & 47.4 \\
30\% & 41.3 & \textbf{35.9} & 72.2 & 48.9 \\
50\% & 43.8 & 35.5 & \textbf{72.9} & \textbf{49.4} \\
100\% (MT) & \textbf{43.9} & 29.1 & 72.6 & 46.1 \\
\bottomrule
\end{tabular}}
\caption{Effects of noise density. We report the average results over different task types and the average results over all the six tasks on the XTREME benchmark. We vary the noise density of the translation span corruption task from 15\% to 100\%. All results are averaged over five runs.}
\label{table:nd}
\end{table}

\subsection{Effects of Noise Density}
In the translation span corruption (TSC) task, the input parallel sentences provide redundant information in two languages, which is different from the standard monolingual span corruption task. Thus, we explore the effects of noise density by varying the noise density in the translation span corruption task, with the other hyperparameters fixed. To reduce the computational load, we do not apply the partially non-autoregressive decoding, i.e., we pretrain the models with the original text-to-text objective. We pretrain \our{} models with the noise density of $0.15$, $0.3$, $0.5$, and $1.0$ respectively. It means $15\%$, $30\%$, $50\%$, or all of the source or target tokens are replaced with the masked tokens. Notice that setting the noise density as $1.0$ is identical to machine translation, where the decoder is required to decode the whole target sentence.

In Table~\ref{table:nd}, we report the average scores on the XTREME benchmark.
From the results, we observe that \our{} achieves the best results with the noise density of $0.5$, rather than a higher noise density such as $1.0$.
The results indicate that the TSC task prefers a higher noise density, so that the model can learn to use more cross-lingual information.
This finding is different from that reported by T5~\cite{t5}, where the span corruption task works better with the noise density of $0.15$ under the monolingual setting.

\section{Related Work}

\paragraph{Cross-lingual LM Pre-training}

Cross-lingual language models are typically built with the Transformer~\cite{transformer} architecture, and pretrained with various pre-training tasks on large-scale text data. 
Multilingual BERT (mBERT;~\citealt{bert}) and XLM-R~\cite{xlmr} are pretrained with masked language modeling (MLM;~\citealt{bert}) on large-scale unlabeled text in about 100 languages. MASS~\cite{mass} and mBART~\cite{mbart} are pretrained in an auto-encoding manner, which provides improvements on the neural machine translation tasks. \mtfive{}~\cite{mt5} is pretrained with the span corruption~\cite{t5} task under the text-to-text formulation~\cite{t5}. Cross-lingual pretrained models also benefit from translation data. XLM~\cite{xlm} jointly learns MLM and the translation language modeling (TLM) task. Unicoder~\cite{unicoder} presents three cross-lingual tasks to learn mappings among languages. 
ALM~\cite{alm} converts the translation pairs into code-switched sequences as the training examples.
Word-aligned BERT models~\cite{Cao2020Multilingual,zhao2020inducing} improves the cross-lingual representations by fine-tuning the mBERT with the objective of minimizing the distance between aligned tokens.
AMBER~\cite{hu2020explicit} propose to maximize the agreement between the forward and backward attention matrices of the input translation pair.
InfoXLM~\cite{infoxlm} proposes the cross-lingual contrastive learning task that maximizes the InfoNCE~\cite{infonce} lower bound of the mutual information between the input translation pair.
XLM-Align~\cite{xlmalign} leverages token-level alignments implied in translation pairs to improve cross-lingual transfer.
XNLG~\cite{xnlg} introduces the cross-lingual transfer for NLG tasks, and achieves zero-shot cross-lingual transfer for question generation and abstractive summarization. VECO~\cite{veco} pretrains a variable cross-lingual pre-training model that learns unified language representations for both NLU and NLG.
ERNIE-M~\cite{erniem} utilizes the back-translation masked language modeling task that generates pseudo parallel sentence pairs for learning TLM.

\paragraph{Encoder-Decoder Pre-training}

\citet{t5} use span corruption to pretrain text-to-text Transformer, where both language understanding and generation tasks are formulated as sequence-to-sequence fine-tuning.
\citet{mass} propose masked sequence-to-sequence pre-training where the model predicts a randomly masked span. 
BART~\cite{bart} design various denoised autoencoding tasks to recover the whole original sentence.
PEGASUS~\cite{pegasus} introduces the gap sentence generation task for abstractive summarization pre-training.
\citet{xnlg} use both denoised autoencoding and machine translation for cross-lingual language generation.
Another strand of research follows unified language model pre-training~\cite{unilm,unilmv2,veco}, where the encoder and the decoder share parameters.
\citet{xlmt,deltalm} reuse pretrained multilingual encoder for sequence-to-sequence pre-training.

\section{Conclusion}

In this paper, we propose \our{} that improves the multilingual text-to-text transfer Transformer with translation data. We introduce three text-to-text pre-training tasks that are built on parallel corpora, and a training objective for improving text-to-text pre-training. Nonetheless, we present a comprehensive comparison of the text-to-text tasks, and show that our \our{} model outperforms \mtfive{} on both cross-lingual understanding and generation benchmarks. For future work, we would like to pretrain \our{} models at a larger scale, and explore more applications, such as machine translation.

\bibliographystyle{acl_natbib}
\bibliography{mt6}

\appendix

\section{Pre-Training Data}

We reconstruct CCNet\footnote{\url{github.com/facebookresearch/cc_net}} and follow~\cite{xlmr} to reproduce the CC-100 corpus for monolingual data. The resulting corpus contains $94$ languages.
We present the language codes and data size in Table~\ref{table:cc} and Table~\ref{table:parallel:data} for the monolingual corpus and parallel corpus, respectively.
Table~\ref{table:cc} reports the language codes and data size in our work. We apply the multilingual sampling strategy~\cite{xlm} with $\alpha=0.7$ for both monolingual and parallel data.

\begin{table}[ht]
\centering
\scriptsize
\begin{tabular}{crcrcr}
\toprule
Code & Size (GB) & Code & Size (GB) & Code & Size (GB) \\ \cmidrule(r){1-2}\cmidrule{3-4}\cmidrule(l){5-6}
af & 0.2 & hr & 1.4 & pa & 0.8 \\
am & 0.4 & hu & 9.5 & pl & 28.6 \\
ar & 16.1 & hy & 0.7 & ps & 0.4 \\
as & 0.1 & id & 17.2 & pt & 39.4 \\
az & 0.8 & is & 0.5 & ro & 11.0 \\
ba & 0.2 & it & 47.2 & ru & 253.3 \\
be & 0.5 & ja & 86.8 & sa & 0.2 \\
bg & 7.0 & ka & 1.0 & sd & 0.2 \\
bn & 5.5 & kk & 0.6 & si & 1.3 \\
ca & 3.0 & km & 0.2 & sk & 13.6 \\
ckb & 0.6 & kn & 0.3 & sl & 6.2 \\
cs & 14.9 & ko & 40.0 & sq & 3.0 \\
cy & 0.4 & ky & 0.5 & sr & 7.2 \\
da & 6.9 & la & 0.3 & sv & 60.4 \\
de & 99.0 & lo & 0.2 & sw & 0.3 \\
el & 13.1 & lt & 2.3 & ta & 7.9 \\
en & 731.6 & lv & 1.3 & te & 2.3 \\
eo & 0.5 & mk & 0.6 & tg & 0.7 \\
es & 85.6 & ml & 1.3 & th & 33.0 \\
et & 1.4 & mn & 0.4 & tl & 1.2 \\
eu & 1.0 & mr & 0.5 & tr & 56.4 \\
fa & 19.0 & ms & 0.7 & tt & 0.6 \\
fi & 5.9 & mt & 0.2 & ug & 0.2 \\
fr & 89.9 & my & 0.4 & uk & 13.4 \\
ga & 0.2 & ne & 0.6 & ur & 3.0 \\
gl & 1.5 & nl & 25.9 & uz & 0.1 \\
gu & 0.3 & nn & 0.4 & vi & 74.5 \\
he & 4.4 & no & 5.5 & yi & 0.3 \\
hi & 5.0 & or & 0.3 & zh & 96.8 \\
\bottomrule
\end{tabular}
\caption{Statistics of CCNet used for pre-training.}
\label{table:cc}
\end{table}

\begin{table}[ht]
\centering
\scriptsize
\begin{tabular}{crcr}
\toprule
ISO Code & Size (GB) & ISO Code & Size (GB) \\ \midrule
en-ar & 5.88 & en-ru & 7.72 \\
en-bg & 0.49 & en-sw & 0.06 \\
en-de & 4.21 & en-th & 0.47 \\
en-el & 2.28 & en-tr & 0.34 \\
en-es & 7.09 & en-ur & 0.39 \\
en-fr & 7.63 & en-vi & 0.86 \\
en-hi & 0.62 & en-zh & 4.02 \\
\bottomrule
\end{tabular}
\caption{Parallel data used for pre-training.}
\label{table:parallel:data}
\end{table}

\section{Hyperparameters for Pre-Training}

As shown in Table~\ref{table:pt-hparam}, we present the hyperparameters for pre-training \our{}.
We extend the vocabulary of the XLM-R~\cite{xlmr} with external 100 unique mask tokens as the vocabulary of \our{} and our \mtfive{} re-implementation.

\begin{table}[ht]
\centering
\small
\renewcommand\tabcolsep{2.8pt}
\begin{tabular}{lr}
\toprule
Hyperparameters & Value \\ \midrule
Layers & 8 \\
Hidden size & 512 \\
FFN inner hidden size & 1,024 \\
Attention heads & 6 \\
Training steps & 500K \\
Batch size & 256 \\
Input length & 512 \\
Adam $\epsilon$ & 1e-6 \\
Adam $\beta$ & (0.9, 0.9999) \\
Learning rate & 1e-4 \\
Learning rate schedule & Linear \\
Warmup steps & 10,000 \\
Gradient clipping & 1.0 \\
Noise density & 0.5 \\
PNAT group number & 3 \\
\bottomrule
\end{tabular}
\caption{Hyperparameters used for pre-training \our{}.}
\label{table:pt-hparam}
\end{table}

\begin{table*}[b]
\centering
\small
\renewcommand\tabcolsep{5.0pt}
\begin{tabular}{lrrrrrrrr}
\toprule
Hyperparameters & WikiAnn & XQuAD & MLQA & TyDiQA & XNLI & PAWS-X & Gigaword & Wikilingua \\ \midrule
Batch size & 32 & 32 & 32 & 32 & 32 & 32 & 32 & 32 \\
Learning rate & 7e-5 & 3e-5 & 3e-5 & 5e-5 & 2e-5 & 3e-5 & 1e-5 & 1e-4 \\
LR schedule & Linear & Linear & Linear & Linear & Linear & Linear & Linear & Linear \\
Warmup & 10\% & 10\% & 10\% & 10\% & 10\% & 10\% & 10K steps & 2.5K steps \\
Epochs/Steps & 5 epochs & 3 epochs & 3 epochs & 40 epochs & 10 epochs & 10 epochs & 20 epochs & 100K steps \\
\bottomrule
\end{tabular}
\caption{Hyperparameters used for fine-tuning \our{} on the end tasks.}
\label{table:hparam}
\end{table*}

\section{Hyperparameters for Fine-Tuning}

In Table~\ref{table:hparam}, we present the hyperparameters for fine-tuning \our{} on the end tasks.

\begin{table*}[b]
\centering
\small
\renewcommand\tabcolsep{4.0pt}
\scalebox{0.85}{
\begin{tabular}{lcccccccccccccccccccc}
\toprule
 Model & ar &   he &   vi &   id &   jv &   ms &   tl &   eu &   ml &   ta &   te &   af &   nl &   en &   de &   el &   bn &   hi &   mr &   ur \\ \midrule
\mtfive{} & 26.5 & 24.0 & 60.7 & 43.5 & 43.7 & 49.2 & 65.2 & 52.4 & 13.1 & 26.4 & 20.2 & 58.2 & 69.4 & 77.5 & 63.6 & 51.7 & 28.3 & 37.9 & 27.2 & 19.6 \\
\our{} & 39.6 & 22.2 & 63.8 & 43.7 & 40.4 & 54.7 & 62.9 & 42.9 & 14.2 & 26.4 & 15.7 & 58.9 & 66.0 & 78.5 & 67.1 & 59.6 & 39.2 & 47.5 & 31.8 & 25.5 \\
\bottomrule
\end{tabular}
}
\renewcommand\tabcolsep{4.0pt}
\scalebox{0.85}{
\begin{tabular}{lccccccccccccccccccccc}
\toprule
 Model & fa &   fr &   it &   pt &   es &   bg &   ru &   ja &   ka &   ko &   th &   sw &   yo &   my &   zh &   kk &   tr &   et &   fi &   hu &  Avg \\ \midrule
\mtfive{} & 15.5 & 69.8 & 69.1 & 67.7 & 57.6 & 61.1 & 49.5 & 24.1 & 26.2 & 23.8 & 3.0 & 54.2 & 56.3 & 2.8 & 29.0 & 23.4 & 52.8 & 57.0 & 62.6 & 60.9 & 43.1 \\
\our{} & 21.7 & 70.7 & 65.9 & 67.8 & 64.9 & 65.8 & 51.6 & 23.4 & 25.3 & 21.9 & 4.9 & 65.2 & 53.6 & 8.5 & 26.3 & 28.6 & 55.9 & 49.3 & 58.2 & 57.1 & 44.7 \\
\bottomrule
\end{tabular}
}
\caption{Results on WikiAnn named entity recognition.}
\label{table:wikiann}
\end{table*}

\begin{table*}[b]
\centering
\small
\renewcommand\tabcolsep{4.0pt}
\scalebox{0.75}{
\begin{tabular}{lcccccccccccc}
\toprule
 Model & en &  es &  de &  el &  ru &  tr &  ar &  vi &  th &  zh &  hi & Avg \\ \midrule
\mtfive{} & 68.6 / 56.7 & 50.2 / 35.6 & 47.2 / 34.1 & 30.3 / 18.5 & 41.4 / 28.5 & 35.9 / 21.9 & 25.1 / 14.7 & 48.6 / 31.6 & 31.7 / 24.6 & 54.7 / 34.9 & 29.7 / 18.6 & 42.1 / 29.1 \\
\our{} & 74.2 / 62.4 & 57.8 / 43.1 & 53.1 / 38.7 & 41.6 / 28.2 & 51.1 / 35.6 & 39.2 / 26.0 & 40.4 / 25.2 & 53.6 / 35.2 & 41.9 / 33.9 & 61.7 / 45.8 & 39.8 / 26.0 & 50.4 / 36.4 \\
\bottomrule
\end{tabular}
}
\caption{Results on XQuAD question answering.}
\label{table:xquad}
\end{table*}

\begin{table*}[b]
\centering
\small
\begin{tabular}{lcccccccc}
\toprule
 Model & en &  es &  de &  ar &  hi &  vi &  zh & Avg \\ \midrule
\mtfive{} & 61.2 / 47.8 & 41.7 / 27.1 & 37.8 / 25.4 & 21.1 / 10.8 & 22.6 / 13.7 & 40.5 / 24.2 & 38.4 / 20.6 & 37.6 / 24.2 \\
\our{} & 65.5 / 52.7 & 47.8 / 32.0 & 43.2 / 29.8 & 32.4 / 18.7 & 31.8 / 20.2 & 45.0 / 28.3 & 42.4 / 23.6 & 44.1 / 29.3 \\
\bottomrule
\end{tabular}
\caption{Results on MLQA question answering.}
\label{table:mlqa}
\end{table*}

\begin{table*}[b]
\centering
\small
\scalebox{0.83}{
\begin{tabular}{lcccccccccc}
\toprule
 Model & en &  ar &  bn &  fi &  id &  ko &  ru &  sw &  te & Avg \\ \midrule
\mtfive{} & 55.4 / 44.7 & 35.3 / 18.3 & 18.4 / 9.2 & 33.3 / 22.2 & 37.3 / 24.8 & 22.6 / 16.9 & 37.3 / 27.7 & 25.5 / 13.6 & 11.2 / 4.5 & 30.7 / 20.2 \\
\our{} & 58.1 / 48.0 & 40.8 / 23.6 & 24.1 / 14.2 & 39.7 / 27.3 & 39.9 / 26.1 & 26.9 / 18.4 & 41.9 / 31.4 & 35.9 / 24.5 & 16.3 / 10.9 & 36.0 / 24.9 \\
\bottomrule
\end{tabular}
}
\caption{Results on TyDiQA question answering.}
\label{table:tydiqa}
\end{table*}

\begin{table*}[b]
\centering
\small
\scalebox{0.95}{
\begin{tabular}{lcccccccccccccccc}
\toprule
 Model & en &  fr &  es &  de &  el &  bg &  ru &  tr &  ar &  vi &  th &  zh &  hi &  sw &  ur & Avg \\ \midrule
\mtfive{} & 75.4 & 62.0 & 62.1 & 58.9 & 58.9 & 57.7 & 59.0 & 55.7 & 52.7 & 58.4 & 55.0 & 55.2 & 53.6 & 42.4 & 50.7 & 57.2 \\
\our{} & 78.4 & 70.6 & 69.8 & 64.8 & 65.7 & 66.6 & 65.8 & 61.6 & 63.3 & 66.6 & 63.1 & 66.2 & 60.3 & 51.5 & 56.9 & 64.7 \\
\bottomrule
\end{tabular}
}
\caption{Results on XNLI natural language inference.}
\label{table:xnli}
\end{table*}

\section{Results on XTREME Cross-Lingual Understanding}

We present the detailed results of the \our{} and our re-implemented \mtfive{} models on XTREME in Table~\ref{table:wikiann}-\ref{table:pawsx}.

\begin{table*}[b]
\centering
\small
\begin{tabular}{lcccccccc}
\toprule
 Model & en &  fr &  de &  es &  ja &  ko &  zh & Avg \\ \midrule
\mtfive{} & 91.6 & 81.2 & 79.9 & 80.7 & 70.7 & 68.2 & 73.5 & 78.0 \\
\our{} & 93.5 & 87.0 & 85.4 & 87.3 & 72.4 & 70.1 & 79.8 & 82.2 \\
\bottomrule
\end{tabular}
\caption{Results on PAWS-X cross-lingual paraphrase adversaries.}
\label{table:pawsx}
\end{table*}

\begin{table*}[t]
\centering
\small
\scalebox{0.98}{
\begin{tabular}{lccc|ccc|ccc|ccc}
\toprule
\multirow{2}{*}{\bf Model} & \multicolumn{3}{c}{\bf es-en} & \multicolumn{3}{c}{\bf ru-en} & \multicolumn{3}{c}{\bf vi-en} & \multicolumn{3}{c}{\bf tr-en} \\
& RG-1 & RG-2 & RG-L & RG-1 & RG-2 & RG-L & RG-1 & RG-2 & RG-L & RG-1 & RG-2 & RG-L \\ \midrule

\mtfive{} & 33.12 & 11.36 & 27.32 & 29.14 &  8.77 & 23.29 & 28.96 &  8.98 & 22.77 & 29.31 & 10.57 & 23.44  \\
\our{} & 33.79 & 11.83 & 27.90 & 30.40 &  9.49 & 24.32 & 29.96 &  9.52 & 23.72 & 29.55 & 10.80 & 23.82  \\
\bottomrule
\end{tabular}
}
\caption{Evaluation results on Wikilingua cross-lingual abstractive summarization. RG is short for ROUGE. Results of \mtfive{} and \our{} are averaged over three runs.}
\label{table:wikilingua-full}
\end{table*}

\section{Results on Wikilingua Cross-Lingual Summarization}

As shown in Table~\ref{table:wikilingua-full}, we present the detailed results of the \our{} and our re-implemented \mtfive{} models on Wikilingua cross-lingual summarization.

\end{document}